\documentclass[11pt,a4paper]{article}
\usepackage[hyperref]{acl2019}
\usepackage{times}
\usepackage{latexsym}

\usepackage{url}

\usepackage{graphicx}
\usepackage{CJK}
\usepackage{multirow}
\usepackage{bm}
\usepackage{subfigure}
\usepackage{amsmath}
\usepackage{amsfonts}
\aclfinalcopy 
\def\aclpaperid{931} 

\title{Chinese Embedding via Stroke and Glyph Information:\\ A Dual-channel View}

\author{Hanqing Tao$^\dag${,\;}
	Shiwei Tong$^\dag${,\;}
	Tong Xu$^{\dag\S}${,\;}
	Qi Liu$^{\dag\S}${,\;}
	Enhong Chen$^{\dag\S}$\thanks{Corresponding author}\\
	$^\dag$Anhui Province Key Laboratory of Big Data Analysis and Application, University of Science\\ and Technology of China \\
	$^\S$School of Data Science, University of Science and Technology of China\\
	\{hqtao, tongsw\}@mail.ustc.edu.cn, \{tongxu, qiliuql, cheneh\}@ustc.edu.cn
}

\date{}

\begin{document}
	\maketitle
	\begin{abstract}		
		
		
		Recent studies have consistently given positive hints that morphology is helpful in enriching word embeddings. In this paper, we argue that Chinese word embeddings can be substantially enriched by the morphological information hidden in characters which is reflected not only in strokes order sequentially, but also in character glyphs spatially. Then, we propose a novel Dual-channel Word Embedding (DWE) model to realize the joint learning of sequential and spatial information of characters. Through the evaluation on both word similarity and word analogy tasks, our model shows its rationality and superiority in modelling the morphology of Chinese.
		
	\end{abstract}		
	
	\section{Introduction}
	Word embeddings are fixed-length vector representations for words \cite{mikolov2013efficient,cui2018survey}.  In recent years, the morphology of words is drawing more and more attention \cite{cotterell2015morphological}, especially for Chinese whose writing system is based on logograms\footnote{https://en.wikipedia.org/wiki/Logogram}.
	
\begin{figure*}[t]
		\centering
		\includegraphics[width=5.5in]{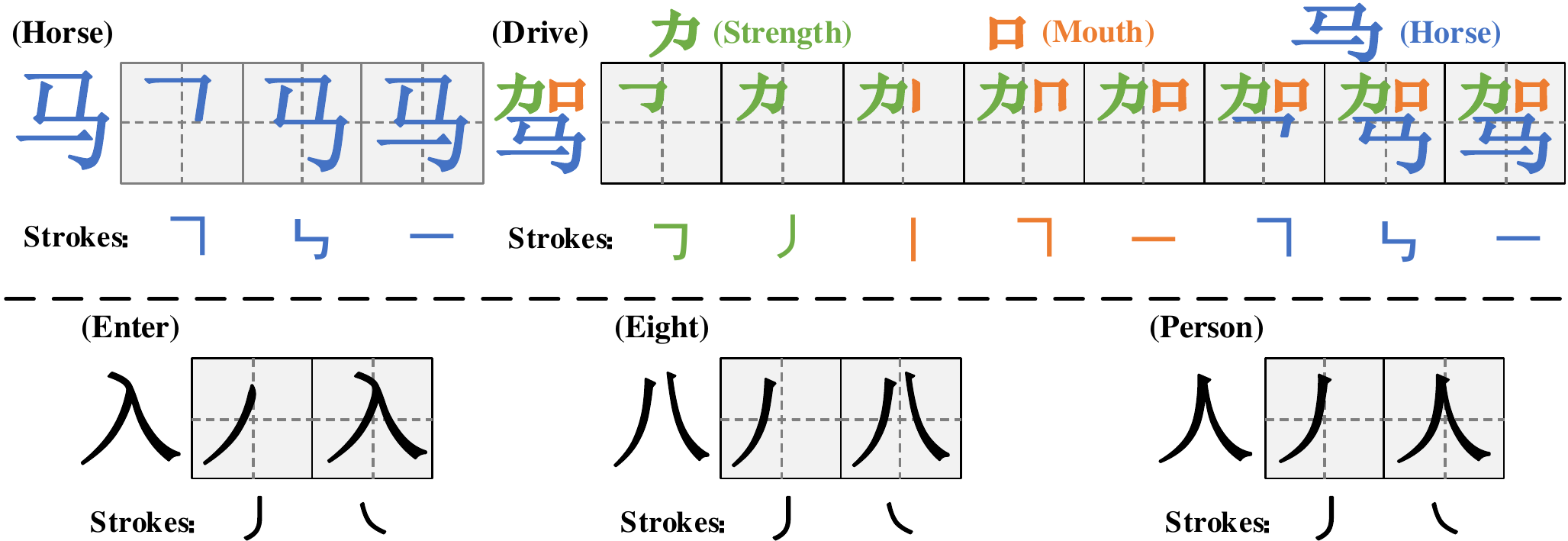}
		\begin{CJK}{UTF8}{gbsn}
			\caption{The upper part is an example for illustrating the inclusion relationship hidden in strokes order and character glyphs. The lower part reflects that a common stroke sequence may form different Chinese characters if their spatial configurations are different.} 
			\label{fig:example}
		\end{CJK}
		\vspace{-0.2cm}
	\end{figure*}

	\begin{CJK}{UTF8}{gbsn}  
		With the gradual exploration of the semantic features of Chinese, scholars have found that not only words and characters are important semantic carriers, but also stroke\footnote{https://en.wikipedia.org/wiki/Stroke\_(CJK\_character)} feature of Chinese characters is crucial for inferring semantics \cite{cao2018cw2vec}. Actually, a Chinese word usually consists of several characters, and each character can be further decomposed into a stroke sequence which is certain and changeless, and this kind of stroke sequence is very similar to the construction of English words. In Chinese, a particular sequence of strokes can reflect the inherent semantics. As shown in the upper half of Figure \ref{fig:example}, the Chinese character ``驾" (drive) can be decomposed into a sequence of eight strokes, where the last three strokes together correspond to a root character ``马" (horse) similar to the root ``clar" of English word ``declare" and ``clarify".
		
		Moreover, Chinese is a language originated from Oracle Bone Inscriptions (a kind of hieroglyphics). Its character glyphs have a spatial structure similar to graphs which can convey abundant semantics \cite{su2017learning}. Additionally, the critical reason why Chinese characters are so rich in morphological information is that they are composed of basic strokes in a 2-D spatial order. However, different spatial configurations of strokes may lead to different semantics. As shown in the lower half of Figure 1, three Chinese characters ``入" (enter), ``八" (eight) and ``人" (man) share exactly a common stroke sequence, but they have completely different semantics because of their different spatial configurations. 
	\end{CJK}
	
	
	In addition, some biological investigations have confirmed that there are actually two processing channels for Chinese language. Specifically, Chinese readers not only activate the left brain which is a dominant hemisphere in processing alphabetic languages \cite{springer1999language,knecht2000language,paulesu2000cultural}, but also activate the areas of the right brain that are responsible for image processing and spatial information at the same time \cite{tan2000brain}. 
	Therefore, we argue that the morphological information of characters in Chinese consists of two parts, i.e., the sequential information hidden in root-like strokes order, and the spatial information hidden in graph-like character glyphs. Along this line, we propose a novel Dual-channel Word Embedding (DWE) model for Chinese to realize the joint learning of sequential and spatial information in characters. Finally, we evaluate DWE on two representative tasks, where the experimental results exactly validate the superiority of DWE in capturing the morphological information of Chinese.

	\section{Releated Work}
	\subsection{Morphological Word Representations}
	Traditional methods on getting word embeddings are mainly based on the distributional hypothesis \cite{harris1954distributional}: words with similar contexts tend to have similar semantics. To explore more interpretable models, some scholars have gradually noticed the importance of the morphology of words in conveying semantics \cite{luong2013better,qiu2014co}, and some studies have proved that the morphology of words can indeed enrich the semantics of word embeddings \cite{sak2010morphology,soricut2015unsupervised,cotterell2015morphological}. More recently, Wieting et al. \shortcite{wieting2016charagram} proposed to represent words using character \textit{n}-gram count vectors. Further, Bojanowski et al. \shortcite{bojanowski2017enriching} improved the classic skip-gram model \cite{mikolov2013efficient} by taking subwords into account in the acquisition of word embeddings, which is instructive for us to regard certain stroke sequences as roots in English. 
	
	
	

	\subsection{Embedding for Chinese Language}
	The complexity of Chinese itself has given birth to a lot of research on Chinese embedding, including the utilization of character features \cite{chen2015joint} and radicals \cite{sun2014radical,yin2016multi,yu2017joint}. Considering the 2-D graphic structure of Chinese characters, Su and Lee \shortcite{su2017learning} creatively proposed to enhance word representations by character glyphs. Lately, Cao et al. \shortcite{cao2018cw2vec} proposed that a Chinese word can be decomposed into a sequence of strokes which correspond to subwords in English, and Wu et al. \shortcite{wu2019glyce} designed a Tianzige-CNN to model the spatial structure of Chinese characters from the perspective of image processing. However, their methods are either somewhat loose for the stroke criteria or unable to capture the interactions between strokes and character glyphs.

	\section{DWE Model}
	As we mentioned earlier, it is reasonable and imperative to learn Chinese word embeddings from two channels, i.e., a sequential stroke \textit{n}-gram channel and a spatial glyph channel. Inspired by the previous works~\cite{chen2015joint, dong2016character, su2017learning, wu2019glyce}, we propose to combine the representation of Chinese words with the representation of characters to obtain finer-grained semantics, so that unknown words can be identified and their relationship with other known Chinese characters can be found by distinguishing the common stroke sequences or character glyph they share. 
	
\begin{figure}[t]
		\centering
		\includegraphics[width=3in]{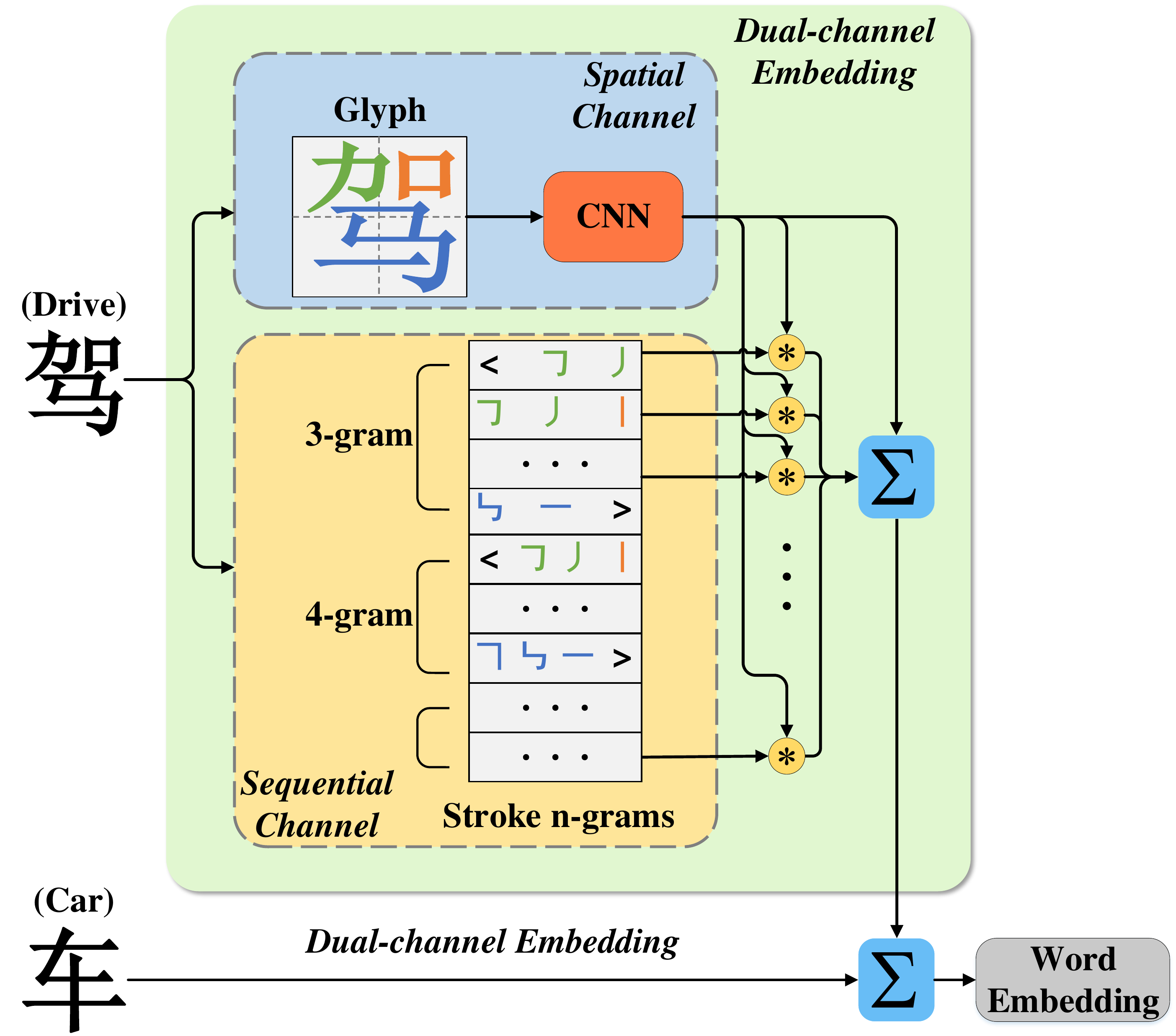}
		\begin{CJK}{UTF8}{gbsn}
			\vspace{-0.4cm}
			\caption{An illustration of our Dual-channel Word Embedding (DWE) model.}
			\label{fig:framework}
		\end{CJK}
		\vspace{-0.4cm}
	\end{figure}

	\begin{CJK}{UTF8}{gbsn}
		Our DWE model is shown in Figure~\ref{fig:framework}. For an arbitrary Chinese word $w$, e.g., ``驾车", it will be firstly decomposed into several characters, e.g., ``驾" and ``车", and each of the characters will be further processed in a dual-channel character embedding sub-module to refine its morphological information. In sequential channel, each character can be decomposed into a stroke sequence according to the criteria of Chinese writing system as shown in Figure~\ref{fig:example}. After retrieving the stroke sequence, we add special boundary symbols $<$ and $>$ at the beginning and end of it and adopt an efficient approach by utilizing the stroke \textit{n}-gram method ~\cite{cao2018cw2vec}\footnote{We apply a richer standard of strokes (32 kinds of strokes) than they did (only 5 kinds of strokes).} to extract strokes order information for each character. More precisely, we firstly scan each character throughout the training corpus and obtain a stroke \textit{n}-gram dictionary $G$. Then, we use $G(c)$ to denote the collection of stroke \textit{n}-grams of each character $c$ in $w$. While in spatial channel, to capture the semantics hidden in glyphs, we render the glyph $I_c$ for each character $c$ and apply a well-known CNN structure, LeNet \cite{lecun1998gradient}, to process each character glyph, which is also helpful to distinguish between those characters that are identical in strokes.
	\end{CJK}
	
	
	After that, we combine the representation of words with the representation of characters and define the word embedding for $w$ as follows:
	\begin{equation}
	\textbf{w} = \textbf{w}_{ID} \oplus \frac{1}{N_c}(\sum_{c \in w} {\sum_{g \in G(c)} \textbf{g} \ast CNN(I_c)}),
	\label{eq:2}
	\end{equation}
	\noindent where $\oplus$ and $\ast$ are compositional operation\footnote{There are a variety of options for $\oplus$ and $\ast$, e.g., addition, item-wise dot product and concatenation. In this paper, we uses the addition operation for $\oplus$ and item-wise dot product operation for $\ast$.}. $\textbf{w}_{ID}$ is the word ID embedding and $N_c$ is the number of characters in $w$.

	According to the previous work~\cite{mikolov2013efficient}, we compute the similarity between current word $w$ and one of its context words $e$ by defining a score function as $s(w, e) = \textbf{w} \cdot \textbf{e}$, where $\textbf{w}$ and $\textbf{e}$ are embedding vectors of $w$ and $e$ respectively. Following the previous works \cite{mikolov2013efficient,bojanowski2017enriching}, the objective function is defined as follows:
	\begin{equation}
	\small
	\begin{split}
	\mathcal{L} = \sum_{w \in D} \sum_{e \in T(w)} log \sigma (s(w, e)) +  \lambda \mathbb{E}_{e'\sim P} [log \sigma(-s(w, e'))],
	\end{split}
	\label{eq:3}
	\end{equation} where $\lambda$ is the number of negative samples and $\mathbb{E}_{e'\sim P}[\cdot]$ is the expectation term. For each $w$ in training corpus $D$, a set of negative samples $T(w)$ will be selected according to the distribution $P$, which is usually set as the word unigram distribution. And $\sigma$ is the sigmoid function.

	

	

	
	\begin{table*}[t]
		\centering
		\footnotesize
		\caption{Performance on word similarity and word analogy task. The dimension of embeddings is set as 300. The evaluation metric is $\rho$ for word similarity and accuracy percentage for word analogy.}
		\begin{tabular}{c|cc|ccc|ccc}
			\hline
			\multirow{3}{*}{Model} & \multicolumn{2}{c|}{\multirow{2}{*}{Word Similarity}} & \multicolumn{6}{c}{Word Analogy} \\
			\cline{4-9}          & \multicolumn{2}{c|}{} & \multicolumn{3}{c|}{3CosAdd} & \multicolumn{3}{c}{3CosMul} \\
			\cline{2-9}          & wordsim-240 & wordsim-296 & Capital & City  & Family & Capital & City  & Family \\
			\hline
			Skipgram \cite{mikolov2013efficient} & 0.5670  & 0.6023  & \textbf{0.7592}  & \textbf{0.8800}  & 0.3676  & \textbf{0.7637}  & \textbf{0.8857}  & 0.3529  \\
			CBOW \cite{mikolov2013efficient}  & 0.5248  & 0.5736  & 0.6499  & 0.6171  & 0.3750  & 0.6219  & 0.5486  & 0.2904  \\
			GloVe \cite{pennington2014glove} & 0.4981  & 0.4019  & 0.6219  & 0.7714  & 0.3167  & 0.5805  & 0.7257  & 0.2375  \\
			sisg \cite{bojanowski2017enriching} & 0.5592  & 0.5884  & 0.4978  & 0.7543  & 0.2610  & 0.5303  & 0.7829  & 0.2206  \\
			\hline
			CWE \cite{chen2015joint}  & 0.5035  & 0.4322  & 0.1846  & 0.1714  & 0.1875  & 0.1713  & 0.1600  & 0.1583  \\
			GWE \cite{su2017learning}  & 0.5531  & 0.5507  & 0.5716  & 0.6629  & 0.2417  & 0.5761  & 0.6914  & 0.2333  \\
			JWE \cite{yu2017joint}  & 0.4734  & 0.5732  & 0.1285  & 0.3657  & 0.2708  & 0.1492  & 0.3771  & 0.2500  \\
			cw2vec \cite{cao2018cw2vec} & 0.5529  & 0.5992  & 0.5081  & 0.7086  & 0.2941  & 0.5465  & 0.7714  & 0.2721  \\
			\hline
			DWE (ours)  & \textbf{0.6105}  & \textbf{0.6137}  & 0.7120  & 0.7486  & \textbf{0.6250}  & 0.6765  & 0.7257  & \textbf{0.6140}  \\
			\hline
		\end{tabular}
		\label{tab:results}%
		\vspace{-0.4cm}
	\end{table*}%

	\section{Experiments}
	
	\subsection{Dataset Preparation}
	We download parts of Chinese Wikipedia articles from Large-Scale Chinese Datasets for NLP\footnote{https://github.com/brightmart/nlp\_chinese\_corpus}. For word segmentation and filtering the stopwords, we apply the jieba\footnote{https://github.com/fxsjy/jieba} toolkit  based on the stopwords table\footnote{https://github.com/YueYongDev/stopwords}. 
	Finally, we get 11,529,432 segmented words.	In accordance with their work~\cite{chen2015joint}, all items whose Unicode falls into the range between 0x4E00 and 0x9FA5 are Chinese characters. We crawl the stroke information of all 20,402 characters from an online dictionary\footnote{https://bihua.51240.com/} and render each character glyph to a 28 $\times$ 28 1-bit grayscale bitmap by using \textit{Pillow}\footnote{https://github.com/python-pillow/Pillow}.  
	
	\subsection{Experimental Setup}
	We choose \textit{adagrad}~\cite{duchi2011adaptive} as our optimizing algorithm, and we set the batch size as 4,096 and learning rate as 0.05. In practice, the slide window size $n$ of stroke $n$-grams is set as $3 \leq n \leq 6$.
	The dimension of all word embeddings of different models is consistently set as 300. We use two test tasks to evaluate the performance of different models: one is \textit{word similarity}, and the other is \textit{word analogy}. A word similarity test consists of multiple word pairs and similarity scores annotated by humans. Good word representations should make the calculated similarity have a high rank correlation with human annotated scores, which is usually measured by the Spearman's correlation $\rho$~\cite{zar1972significance}.
	
	The form of an analogy problem is like ``king":``queen" = ``man":``?", and ``woman" is the most proper answer to ``?". That is, in this task, given three words $a$, $b$, and $h$, the goal is to infer the fourth word $t$ which satisfies ``$a$ is to $b$ that is similar to $h$ is to $t$". We use $3CosAdd$ \cite{mikolov2013efficient} and $3CosMul$ function \cite{levy2014linguistic} to calculate the most appropriate word $t$. By using the same data used in \cite{chen2015joint} and \cite{cao2018cw2vec}, we adopt two manually-annotated datasets for Chinese word similarity task, i.e., wordsim-240 and wordsim-296~\cite{jin2012semeval} and a three-group\footnote{capitals of countries, (China) states/provinces of cities, and family relations. } dataset for Chinese word analogy task.

	\subsection{Baseline Methods}
	We use gensim\footnote{https://radimrehurek.com/gensim/} to implement both CBOW and Skipgram and apply the source codes pulished by the authors to implement CWE\footnote{https://github.com/Leonard-Xu/CWE}, JWE\footnote{https://github.com/hkust-knowcomp/jwe}, GWE\footnote{https://github.com/ray1007/GWE} and GloVe\footnote{https://github.com/stanfordnlp/GloVe}. Since Cao et al.~\shortcite{cao2018cw2vec} did not publish their code, we follow their paper and reproduce cw2vec in mxnet\footnote{https://mxnet.apache.org/} which we also use to implement sisg~\cite{bojanowski2017enriching}\footnote{ http://gluon-nlp.mxnet.io/master/examples/word\_embed-ding/word\_embedding\_training.html} and our DWE. To encourage further research, we will publish our model and datasets.

	\subsection{Experimental Results}
	\begin{CJK}{UTF8}{gbsn}
		The experimental results are shown in Table \ref{tab:results}. We can observe that our DWE model achieves the best results both on dataset wordsim-240 and wordsim-296 in the similarity task as expected because of the particularity of Chinese morphology, but it only improves the accuracy for the \textit{family} group in the analogy task. 
		
		Actually, it is not by chance that we get these results, because DWE has the advantage of distinguishing between morphologically related words, which can be verified by the results of the similarity task. Meanwhile, in the word analogy task, those words expressing family relations in Chinese are mostly compositional in their character glyphs. For example, in an analogy pair ``兄弟" (brother) : ``姐妹" (sister) = ``儿子" (son) : ``女儿" (daughter), we can easily find that ``兄弟" and ``儿子" share an exactly common part of glyph ``儿" (male relative of a junior generation) while ``姐妹" and ``女儿" share an exactly common part of glyph ``女" (female), and this kind of morphological pattern can be accurately captured by our model. However, most of the names of countries, capitals and cities are transliterated words, and the relationship between the morphology and semantics of words is minimal, which is consistent with the findings reported in  ~\cite{su2017learning}. For instance, in an analogy pair ``西班牙" (Spain) : ``马德里" (Madrid) = ``法国" (France) : ``巴黎" (Paris), we cannot infer any relevance among these four words literally because they are all translated by pronunciation.
		
		
		In summary, since different words that are morphologically similar tend to have similar semantics in Chinese, simultaneously modeling the sequential and spatial information of characters from both stroke \textit{n}-grams and glyph features can indeed improve the modeling of Chinese word representations substantially.
	\end{CJK}

	\section{Conclusions}
	In this article, we first analyzed the similarities and differences in terms of morphology between alphabetical languages and Chinese. Then, we delved deeper into the particularity of Chinese morphology and proposed our DWE model by taking into account the sequential information of strokes order and the spatial information of glyphs. Through the evaluation on two representative tasks, our model shows its superiority in capturing the morphological information of Chinese.

	\bibliography{reference}

\begin{thebibliography}{28}
\expandafter\ifx\csname natexlab\endcsname\relax\def\natexlab#1{#1}\fi

\bibitem[{Bojanowski et~al.(2017)Bojanowski, Grave, Joulin, and
  Mikolov}]{bojanowski2017enriching}
Piotr Bojanowski, Edouard Grave, Armand Joulin, and Tomas Mikolov. 2017.
\newblock Enriching word vectors with subword information.
\newblock \emph{Transactions of the Association for Computational Linguistics},
  5:135--146.

\bibitem[{Cao et~al.(2018)Cao, Lu, Zhou, and Li}]{cao2018cw2vec}
Shaosheng Cao, Wei Lu, Jun Zhou, and Xiaolong Li. 2018.
\newblock cw2vec: Learning chinese word embeddings with stroke n-gram
  information.
\newblock In \emph{Thirty-Second AAAI Conference on Artificial Intelligence}.

\bibitem[{Chen et~al.(2015)Chen, Xu, Liu, Sun, and Luan}]{chen2015joint}
Xinxiong Chen, Lei Xu, Zhiyuan Liu, Maosong Sun, and Huanbo Luan. 2015.
\newblock Joint learning of character and word embeddings.
\newblock In \emph{Twenty-Fourth International Joint Conference on Artificial
  Intelligence}.

\bibitem[{Cotterell and Sch{\"u}tze(2015)}]{cotterell2015morphological}
Ryan Cotterell and Hinrich Sch{\"u}tze. 2015.
\newblock Morphological word-embeddings.
\newblock In \emph{Proceedings of the 2015 Conference of the North American
  Chapter of the Association for Computational Linguistics: Human Language
  Technologies}, pages 1287--1292.

\bibitem[{Cui et~al.(2018)Cui, Wang, Pei, and Zhu}]{cui2018survey}
Peng Cui, Xiao Wang, Jian Pei, and Wenwu Zhu. 2018.
\newblock A survey on network embedding.
\newblock \emph{IEEE Transactions on Knowledge and Data Engineering}.

\bibitem[{Dong et~al.(2016)Dong, Zhang, Zong, Hattori, and
  Di}]{dong2016character}
Chuanhai Dong, Jiajun Zhang, Chengqing Zong, Masanori Hattori, and Hui Di.
  2016.
\newblock Character-based lstm-crf with radical-level features for chinese
  named entity recognition.
\newblock In \emph{Natural Language Understanding and Intelligent
  Applications}, pages 239--250. Springer.

\bibitem[{Duchi et~al.(2011)Duchi, Hazan, and Singer}]{duchi2011adaptive}
John Duchi, Elad Hazan, and Yoram Singer. 2011.
\newblock Adaptive subgradient methods for online learning and stochastic
  optimization.
\newblock \emph{Journal of Machine Learning Research}, 12(Jul):2121--2159.

\bibitem[{Harris(1954)}]{harris1954distributional}
Zellig~S Harris. 1954.
\newblock Distributional structure.
\newblock \emph{Word}, 10(2-3):146--162.

\bibitem[{Jin and Wu(2012)}]{jin2012semeval}
Peng Jin and Yunfang Wu. 2012.
\newblock Semeval-2012 task 4: evaluating chinese word similarity.
\newblock In \emph{Proceedings of the First Joint Conference on Lexical and
  Computational Semantics-Volume 1: Proceedings of the main conference and the
  shared task, and Volume 2: Proceedings of the Sixth International Workshop on
  Semantic Evaluation}, pages 374--377. Association for Computational
  Linguistics.

\bibitem[{Knecht et~al.(2000)Knecht, Deppe, Dr{\"a}ger, Bobe, Lohmann,
  Ringelstein, and Henningsen}]{knecht2000language}
Stefan Knecht, Michael Deppe, Bianca Dr{\"a}ger, L~Bobe, Hubertus Lohmann, E-B
  Ringelstein, and H~Henningsen. 2000.
\newblock Language lateralization in healthy right-handers.
\newblock \emph{Brain}, 123(1):74--81.

\bibitem[{LeCun et~al.(1998)LeCun, Bottou, Bengio, Haffner
  et~al.}]{lecun1998gradient}
Yann LeCun, L{\'e}on Bottou, Yoshua Bengio, Patrick Haffner, et~al. 1998.
\newblock Gradient-based learning applied to document recognition.
\newblock \emph{Proceedings of the IEEE}, 86(11):2278--2324.

\bibitem[{Levy and Goldberg(2014)}]{levy2014linguistic}
Omer Levy and Yoav Goldberg. 2014.
\newblock Linguistic regularities in sparse and explicit word representations.
\newblock In \emph{Proceedings of the eighteenth conference on computational
  natural language learning}, pages 171--180.

\bibitem[{Luong et~al.(2013)Luong, Socher, and Manning}]{luong2013better}
Thang Luong, Richard Socher, and Christopher Manning. 2013.
\newblock Better word representations with recursive neural networks for
  morphology.
\newblock In \emph{Proceedings of the Seventeenth Conference on Computational
  Natural Language Learning}, pages 104--113.

\bibitem[{Mikolov et~al.(2013)Mikolov, Chen, Corrado, and
  Dean}]{mikolov2013efficient}
Tomas Mikolov, Kai Chen, Greg Corrado, and Jeffrey Dean. 2013.
\newblock Efficient estimation of word representations in vector space.
\newblock \emph{arXiv preprint arXiv:1301.3781}.

\bibitem[{Paulesu et~al.(2000)Paulesu, McCrory, Fazio, Menoncello, Brunswick,
  Cappa, Cotelli, Cossu, Corte, Lorusso et~al.}]{paulesu2000cultural}
Eraldo Paulesu, Eamon McCrory, Ferruccio Fazio, Lorena Menoncello, Nicola
  Brunswick, Stefano~F Cappa, Maria Cotelli, Giuseppe Cossu, Francesca Corte,
  M~Lorusso, et~al. 2000.
\newblock A cultural effect on brain function.
\newblock \emph{Nature neuroscience}, 3(1):91.

\bibitem[{Pennington et~al.(2014)Pennington, Socher, and
  Manning}]{pennington2014glove}
Jeffrey Pennington, Richard Socher, and Christopher Manning. 2014.
\newblock Glove: Global vectors for word representation.
\newblock In \emph{Proceedings of the 2014 conference on empirical methods in
  natural language processing (EMNLP)}, pages 1532--1543.

\bibitem[{Qiu et~al.(2014)Qiu, Cui, Bian, Gao, and Liu}]{qiu2014co}
Siyu Qiu, Qing Cui, Jiang Bian, Bin Gao, and Tie-Yan Liu. 2014.
\newblock Co-learning of word representations and morpheme representations.
\newblock In \emph{Proceedings of COLING 2014, the 25th International
  Conference on Computational Linguistics: Technical Papers}, pages 141--150.

\bibitem[{Sak et~al.(2010)Sak, Saraclar, and
  G{\"u}ng{\"o}r}]{sak2010morphology}
Ha{\c{s}}im Sak, Murat Saraclar, and Tunga G{\"u}ng{\"o}r. 2010.
\newblock Morphology-based and sub-word language modeling for turkish speech
  recognition.
\newblock In \emph{2010 IEEE International Conference on Acoustics, Speech and
  Signal Processing}, pages 5402--5405. IEEE.

\bibitem[{Soricut and Och(2015)}]{soricut2015unsupervised}
Radu Soricut and Franz Och. 2015.
\newblock Unsupervised morphology induction using word embeddings.
\newblock In \emph{Proceedings of the 2015 Conference of the North American
  Chapter of the Association for Computational Linguistics: Human Language
  Technologies}, pages 1627--1637.

\bibitem[{Springer et~al.(1999)Springer, Binder, Hammeke, Swanson, Frost,
  Bellgowan, Brewer, Perry, Morris, and Mueller}]{springer1999language}
Jane~A Springer, Jeffrey~R Binder, Thomas~A Hammeke, Sara~J Swanson, Julie~A
  Frost, Patrick~SF Bellgowan, Cameron~C Brewer, Holly~M Perry, George~L
  Morris, and Wade~M Mueller. 1999.
\newblock Language dominance in neurologically normal and epilepsy subjects: a
  functional mri study.
\newblock \emph{Brain}, 122(11):2033--2046.

\bibitem[{Su and Lee(2017)}]{su2017learning}
Tzu-Ray Su and Hung-Yi Lee. 2017.
\newblock Learning chinese word representations from glyphs of characters.
\newblock \emph{arXiv preprint arXiv:1708.04755}.

\bibitem[{Sun et~al.(2014)Sun, Lin, Yang, Ji, and Wang}]{sun2014radical}
Yaming Sun, Lei Lin, Nan Yang, Zhenzhou Ji, and Xiaolong Wang. 2014.
\newblock Radical-enhanced chinese character embedding.
\newblock In \emph{International Conference on Neural Information Processing},
  pages 279--286. Springer.

\bibitem[{Tan et~al.(2000)Tan, Spinks, Gao, Liu, Perfetti, Xiong, Stofer, Pu,
  Liu, and Fox}]{tan2000brain}
Li~Hai Tan, John~A Spinks, Jia-Hong Gao, Ho-Ling Liu, Charles~A Perfetti, Jinhu
  Xiong, Kathryn~A Stofer, Yonglin Pu, Yijun Liu, and Peter~T Fox. 2000.
\newblock Brain activation in the processing of chinese characters and words: a
  functional mri study.
\newblock \emph{Human brain mapping}, 10(1):16--27.

\bibitem[{Wieting et~al.(2016)Wieting, Bansal, Gimpel, and
  Livescu}]{wieting2016charagram}
John Wieting, Mohit Bansal, Kevin Gimpel, and Karen Livescu. 2016.
\newblock Charagram: Embedding words and sentences via character n-grams.
\newblock \emph{arXiv preprint arXiv:1607.02789}.

\bibitem[{Wu et~al.(2019)Wu, Meng, Han, Li, Li, Mei, Nie, Sun, and
  Li}]{wu2019glyce}
Wei Wu, Yuxian Meng, Qinghong Han, Muyu Li, Xiaoya Li, Jie Mei, Ping Nie,
  Xiaofei Sun, and Jiwei Li. 2019.
\newblock Glyce: Glyph-vectors for chinese character representations.
\newblock \emph{arXiv preprint arXiv:1901.10125}.

\bibitem[{Yin et~al.(2016)Yin, Wang, Li, Li, and Wang}]{yin2016multi}
Rongchao Yin, Quan Wang, Peng Li, Rui Li, and Bin Wang. 2016.
\newblock Multi-granularity chinese word embedding.
\newblock In \emph{Proceedings of the 2016 Conference on Empirical Methods in
  Natural Language Processing}, pages 981--986.

\bibitem[{Yu et~al.(2017)Yu, Jian, Xin, and Song}]{yu2017joint}
Jinxing Yu, Xun Jian, Hao Xin, and Yangqiu Song. 2017.
\newblock Joint embeddings of chinese words, characters, and fine-grained
  subcharacter components.
\newblock In \emph{Proceedings of the 2017 Conference on Empirical Methods in
  Natural Language Processing}, pages 286--291.

\bibitem[{Zar(1972)}]{zar1972significance}
Jerrold~H Zar. 1972.
\newblock Significance testing of the spearman rank correlation coefficient.
\newblock \emph{Journal of the American Statistical Association},
  67(339):578--580.

\end{thebibliography}
	\bibliographystyle{acl_natbib}
	
\end{document}